\documentclass[runningheads]{llncs}
\usepackage{graphicx}
\usepackage{comment}
\usepackage{tabularx}
\usepackage{epsfig}
\usepackage{graphicx}
\usepackage{amsmath}
\usepackage{amssymb}
\usepackage{bm}
\usepackage{tabularx}
\usepackage{float}
\usepackage{setspace}
\usepackage{booktabs}
\usepackage[colorlinks,
            linkcolor=red,
            anchorcolor=blue,
            citecolor=green
            ]{hyperref}
\usepackage{graphicx}
\usepackage{setspace}
\usepackage{amsmath,amssymb} % define this before the line numbering.
\usepackage{color}
\begin{document}
% \linenumbers
\pagestyle{headings}
\mainmatter
\def\ECCVSubNumber{2638}  % Insert your submission number here

\title{Class-wise Dynamic Graph Convolution for Semantic Segmentation} % Replace with your title

\begin{comment}
\titlerunning{ECCV-20 submission ID \ECCVSubNumber} 
\authorrunning{ECCV-20 submission ID \ECCVSubNumber} 
\author{Anonymous ECCV submission}
\institute{Paper ID \ECCVSubNumber}
\end{comment}
%******************

%\begin{comment}
\titlerunning{Class-wise Dynamic Graph Convolution}
\author{Hanzhe Hu\inst{1}\orcidID{0000-0003-2799-2655}* \and
Deyi Ji\inst{2} \orcidID{0000-0001-7561-9789} \and \\
Weihao Gan\inst{2}  \and 
Shuai Bai \inst{3}  \and 
Wei Wu \inst{2} \and
Junjie Yan \inst{2}}
\authorrunning{H. Hu et al.}
\institute{Peking University, Beijing, China \and
SenseTime Group Limited, Beijing, China \and
Beijing University of Posts and Telecommunications, Beijing, China\\
\email{huhz@pku.edu.cn, \{jideyi,ganweihao,wuwei,\\yanjunjie\}@sensetime.com, baishuai@bupt.edu.cn}}

%\end{comment}
%******************
\maketitle

\begin{abstract}
Recent works have made great progress in semantic segmentation by exploiting contextual information in a local or global manner with dilated convolutions, pyramid pooling or self-attention mechanism. In order to avoid potential misleading contextual information aggregation in previous works, we propose a class-wise dynamic graph convolution(CDGC) module to adaptively propagate information. The graph reasoning is performed among pixels in the same class. Based on the proposed CDGC module, we further introduce the Class-wise Dynamic Graph Convolution Network(CDGCNet), which consists of two main parts including the CDGC module and a basic segmentation network, forming a coarse-to-fine paradigm. Specifically, the CDGC module takes the coarse segmentation result as class mask to extract node features for graph construction and performs dynamic graph convolutions on the constructed graph to learn the feature aggregation and weight allocation. Then the refined feature and the original feature are fused to get the final prediction. We conduct extensive experiments on three popular semantic segmentation benchmarks including Cityscapes, PASCAL VOC 2012 and COCO Stuff, and achieve state-of-the-art performance on all three benchmarks. 
%\dots
\end{abstract}

\section{Introduction}

\renewcommand{\thefootnote}{\fnsymbol{footnote}}
\footnotetext[0]{*\ This work is done when Hanzhe Hu is an intern at SenseTime Group Limited.}

Semantic Segmentation is a fundamental and challenging problem in computer vision, which aims to 
assign a category label to each pixel in an image. It has been widely  applied to many scenarios, 
such as autonomous driving, scene understanding and image editing. 

Recent  state-of-the-art semantic segmentation methods based on the fully convolutional network(FCN)~\cite{long2015fully} have made great progress. 
%However, due to the limited size of the receptive fields, the performance of FCN-based method is strongly restricted with insufficient long-range contextual information. 
To capture the long-range contextual information, the atrous spatial pyramid pooling(ASPP) module in DeepLabv3~\cite{chen2017rethinking} aggregates spatial regularly sampled pixels at different dilated rates and the pyramid pooling module in PSPNet~\cite{zhao2017pyramid} partitions the feature maps into multiple regions before pooling. %, which means the gathered context is a mixture of pixels that might belong to different categories, leading to a low-reliability label prediction. 
More comprehensively, PSANet~\cite{zhao2018psanet} was proposed to generate dense and pixel-wise contextual information, which learns to aggregate information via  a predicted attention map. Non-local Network~\cite{wang2018non} adopts self-attention mechanism, which enables every pixel to receive information from every other pixels in the image, resulting in a much complete pixel-wise representation. 
\begin{figure}
    \centering
    \includegraphics[height=4cm]{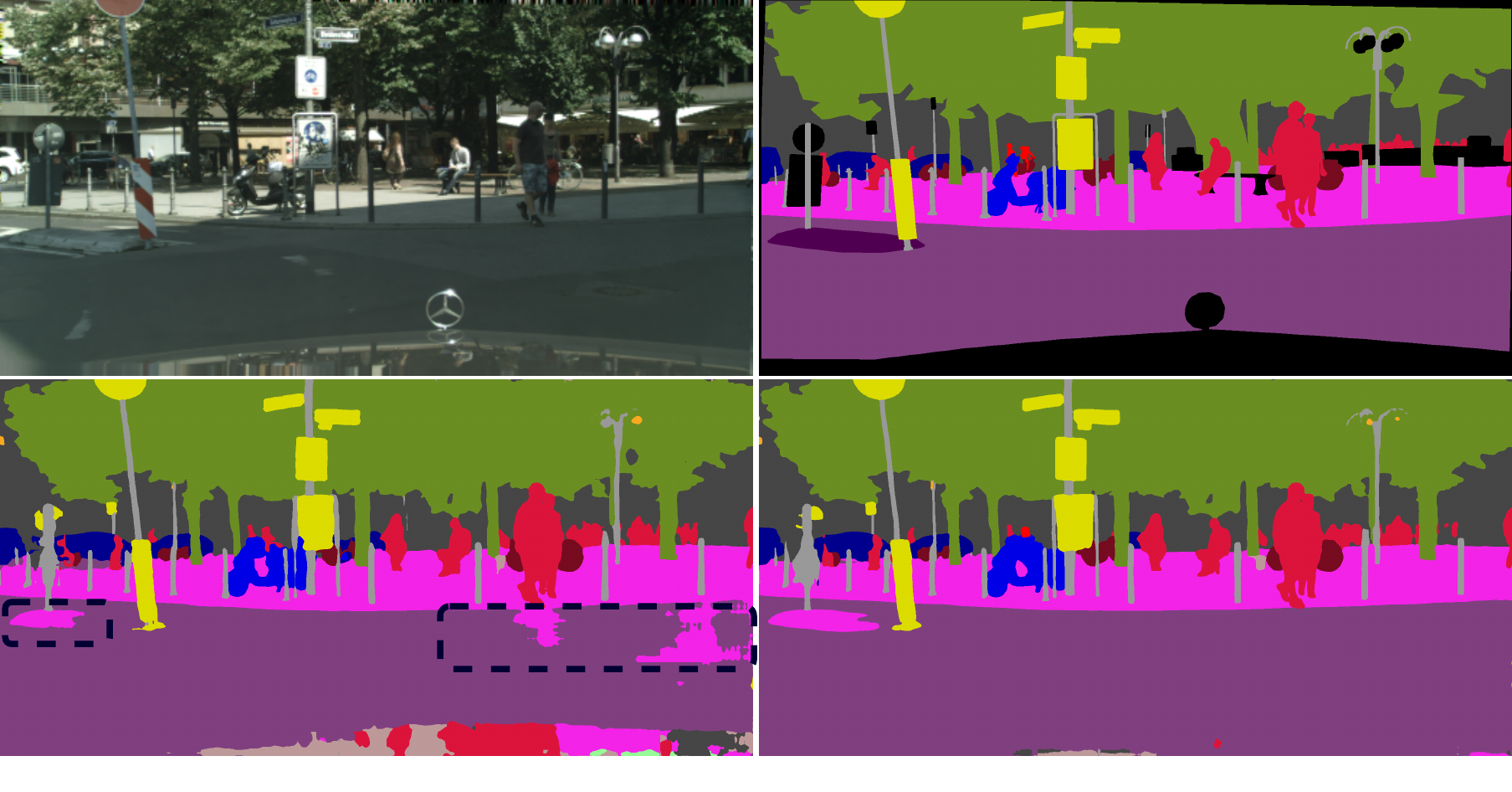}
    \caption{Viusal example from left to right, top to bottom is : original image, groundtruth, deeplabv3 result, the proposed CDGCNet result. From the two indicated regions, our method preserves more contextual details and accurate prediction along boundaries.}
    \label{fig:intro}
\end{figure}

However, the ways of utilizing the contextual information in existing approaches are still problematic. From one point of view, larger receptive field in deeper network is necessary for semantic prediction. Also, dilated based or the pooling based methods take even larger contextual information into consideration. These two operations are neither adaptive nor friendly to pixel-wised segmentation prediction problem. Another view of self-attention based methods (PSANet~\cite{zhao2018psanet}, Non-local Network~\cite{wang2018non}, and etc~\cite{fu2019dual,huang2018ccnet,yuan2018ocnet,Li_2019_ICCV}) is that, pixels from long-range non-local regions have different feature representations, which results in major issues on two aspects when optimizing the convolution neural network. First, contextual information is learned from previous network layers by considering the local and non-local cues. Considering the large variations and uncorrelations in contextual representations, weighted convoluting all the regions together results in difficulties of learning discriminative pixel-level features. For example, feature of a sky location with neighborhood tree region should be different from the one of a sky location with building region, which should not be learned together. Second, contextual information is also class-specific. That means, feature of a tree region is not proper to contribute to the learning of a sky region. The target is to directly distinguish whether the region is a sky region or not. 

Aiming to address the above issues, we propose the Class-wise Dynamic Graph Convolution Network (CDGCNet), which can efficiently utilize the long-range contextual dependencies and aggregate the useful information for better pixel label prediction. Since graph convolution is remarkable at leveraging relations between nodes and can serve as a suitable reasoning method. It is worth noting that self-attention method is actually to build a fully-connected graph, so we further improve the structure of plain GCN for better performance. First, we adopt the class-wise strategy to construct the graph (node and edge) for each class, so that the useful information for each class can be independently learned. Second, for the graph of each class, not all the context regions are included during graph reasoning. Specifically, the hard positive and negative regions are dynamically identified into the graph transform. With these two designs in graph, the most important contextual information can be exploited for pixel level semantic prediction.

The overall framework of the proposed CDGCNet method is shown in Fig.~\ref{fig:framework}, which follows the coarse-to-fine paradigm. The first part is a simple but complete semantic segmentation network, called basic network, which can generate coarse prediction map and it can be any of  state-of-the-art semantic segmentation architectures. The second part is the CDGC module. Firstly, the CDGC module takes coarse prediction map and feature map from the basic network as inputs, and transforms the prediction map into class mask to extract node features from different classes for graph construction. After that, for each class, dynamic graph convolution is performed on the constructed graph to learn the feature aggregation and weight allocation. Finally, the refined feature and the original feature are fused to get the final prediction.

The main contributions of this paper are summarized as follows:
\begin{itemize}
    \item[$\bullet$] The proposed CDGCNet utilizes a class-wise learning strategy so that semantically related features are considered for contextual learning.
     \item[$\bullet$] During the graph construction on each class, hard positive and hard negative information are dynamically sampled from the coarse segmentation result, which avoids heavy graph connections  and benefits the feature learning.
    %It not only , but also improves the heavy computation cost. 
    \item[$\bullet$] We conduct extensive experiments on several public datasets, and obtain state-of-the-art performances on the Cityscapes~\cite{cordts2016cityscapes}, PASCAL VOC 2012~\cite{everingham2010pascal} and COCO Stuff~\cite{caesar2018coco} datasets.
 \end{itemize}

%-------------------------------------------------------------------------

\section{Related Work}
\noindent
\textbf{Semantic Segmentation. } Benefiting from the success of deep neural networks~\cite{krizhevsky2012imagenet,simonyan2014very,he2016deep}, semantic segmentation has achieved great progress. FCN~\cite{long2015fully} is the first approach to adopt fully convolutional network for semantic segmentation. Later, many FCN-baed works are proposed, such as UNet~\cite{ronneberger2015u}, SegNet~\cite{badrinarayanan2017segnet}, RefineNet~\cite{lin2017refinenet}, PSPNet~\cite{zhao2017pyramid}, DeepLab series~\cite{chen2014semantic,chen2017deeplab,chen2017rethinking,chen2018encoder}. Chen \textit{et al.}~\cite{chen2017deeplab} and Yu \textit{et al.}~\cite{yu2015multi} removed the last two downsample layers to obtain a dense prediction and utilized dilated convolutions to enlarge the receptive field. In our model, we also adopt the above paradigm to get a better feature map and hence, improve the performance of the model. 

\noindent
\textbf{Context. } Context plays a critical role in various vision tasks including semantic segmentation. Many works are proposed to generate better feature representations by exploiting better contextual information. From the spatial perspective, DeepLab v3~\cite{chen2017rethinking} employs multiple atrous convolutions with different dilation rates to capture contextual information, while PSPNet~\cite{zhao2017pyramid} employs pyramid pooling over sub-regions of four pyramid scales to harvest information. These methods, however, are all focusing on enlarging receptive fields in a local perspective and hence lose global context information. While from the attention perspective, Wang \textit{et al.}~\cite{wang2018non} extend the idea of self-attention from transformer~\cite{vaswani2017attention} into the vision field and proposed the non-local module to generate the attention map by calculating the correlation matrix between each spatial point in the feature map, and then the attention map guides the dense contextual information aggregation. Later, DANet~\cite{fu2019dual} applied both spatial and channel attention to gather information around the feature maps
%, PSANet~\cite{zhao2018psanet} constructs the pixel-wise attention map based on each pixel independently
. Unlike works mentioned above, our proposed module separately allocates attention to pixels belonging to the same category, effectively avoiding wrong contextual information aggregation. 

\noindent
\textbf{Graph Reasoning. } Graph-based methods have been very popular these days and shown to be an efficient way of relation reasoning. CRF~\cite{chandra2017dense} is proposed based on the graph model for image segmentation and works as an effective postprocessing method in DeepLab~\cite{chen2017deeplab}. Recently, Graph Convolution Networks(GCN)~\cite{kipf2016semi} are proposed for semi-supervised classification, and Wang \textit{et al.}~\cite{wang2018videos} use GCN to capture relations between objects in video recognition tasks. Later, a few works based on GCN have been proposed onto the semantic segmentation problem, including~\cite{chen2019graph,liang2018symbolic,li2018beyond}, which all similarly model the relations between regions of the image rather than individual pixels. Concretely, clusters of pixels are defined as the vertices of the graph, hence graph reasoning is performed in the intermediate space projected from the original feature space to reduce computation cost%, thus losing some ratio of information
. Different from these recent GCN-based methods, we perform graph convolution in a class-wise manner, where GCNs are employed only to the nodes in the same category, leading to a better feature learning. The refined features thus can provide a better prediction result in semantic segmentation task.

%-------------------------------------------------------------------------

\section{Approach}

%-------------------------------------------------------------------------
In this section, we will describe the proposed class-wise dynamic graph convolution (CDGC) module in detail. Firstly, we will revisit the basic knowledge of graph convolutional network. Then we will present a general framework of our network and introduce class-wise dynamic graph convolution module which separately performs graph reasoning on the pixels within the same category, hence producing a refined prediction map for semantic segmentation. Finally, we will bring out the supervision manner of the proposed model.

%-------------------------------------------------------------------------
\subsection{Preliminaries}

\noindent
\textbf{Graph Convolution.} Given an input feature $X\in \mathbb{R}^{N\times D}$, where $N$ is the number of nodes in the feature map and D is the feature dimension, we can build a feature graph G from this input feature. Specifically, the graph G can be formulated as
$G = \{ \textit{V, $\varepsilon$, A} \}$ 
with $\textit{V}$ as its nodes, \textit{$\varepsilon$} as its edges and $\textit{A}$ as its adjacency matrix. Normally, the adjacency matrix A is a binary matrix, in practice, we try many ways to construct the graph, including top-k binary matrix or dynamic learnable matrix, and further design a novel dynamic sampling method to construct the graph and perform extensive experiments to verify its validity. Intuitively, unlike standard convolutions which operates on a local regular grid, the graph enables us to compute the response of a node based on its neighbors defined in the adjacency matrix, hence receiving a much wider receptive field than regular convolutions. Formally, the graph convolution is defined as,
\begin{equation}
    Z = \sigma(AXW) ,
\end{equation}
where $\sigma(\cdot)$ denotes the non-linear activation function, $A\in \mathbb{R}^{N\times N}$ is the adjacency matrix measuring the relations of nodes in the graph and $W\in \mathbb{R}^{D\times D}$ is the weight matrix. In our experiments, we use ReLU as activation function and perform experiments with different graph construction methods.

\begin{figure*}[!h]
    \centering
    \includegraphics[width=1\textwidth]{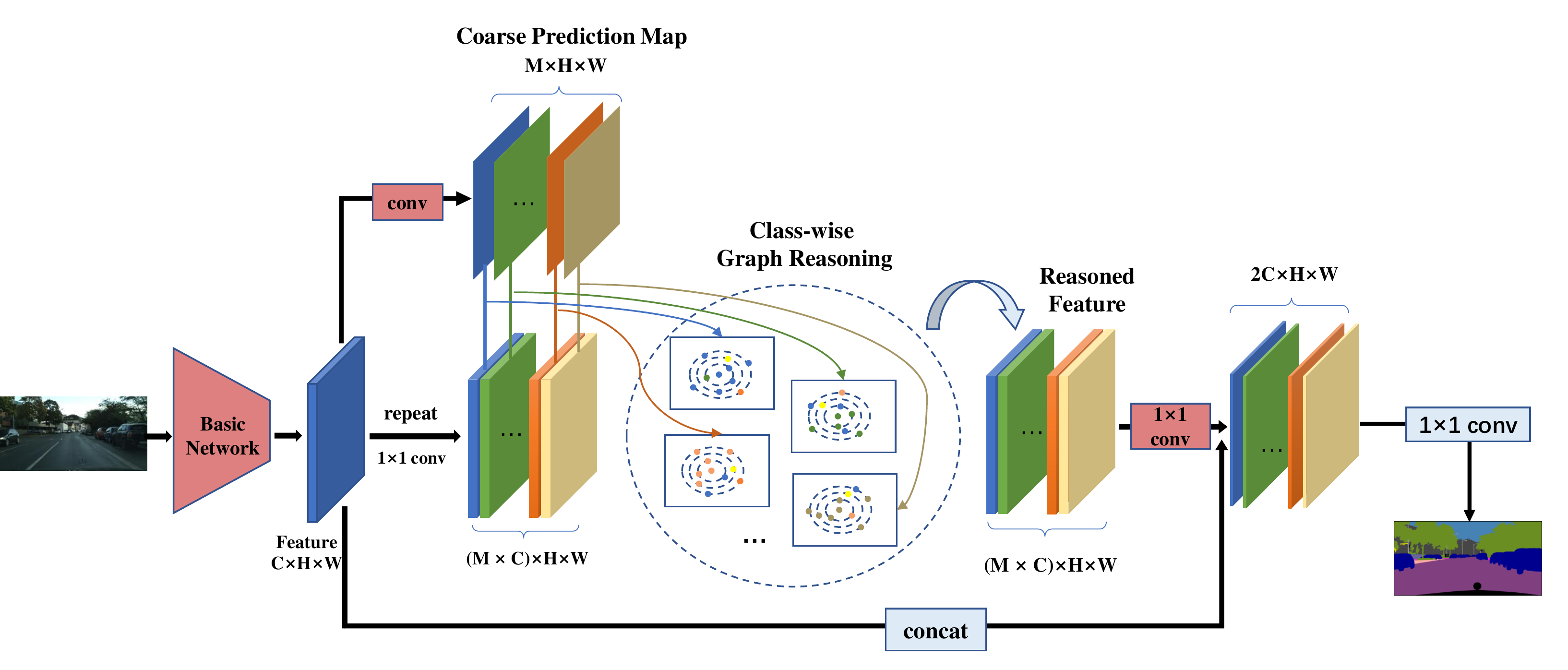}
    \caption{An Overview of the Class-wise Dynamic Graph Convolution Network. Given an input image, we first feed it into the basic segmentation network to get the high-level feature map and the corresponding coarse segmentation result. Then the CDGC module is applied to preform graph reasoning along nodes of the feature map, producing a refined feature which is subsequently fused with the original feature to get the final refined segmentation result. Specially, in the class-wise graph reasoning part, different colors of lines and dots denote different classes of pixels, under the guidance of coarse prediction map, most positive pixels are sampled while also harvesting few hard pixels in different colors from the target color. }
    \label{fig:framework}
\end{figure*}

\subsection{Overall Framework}

As illustrated in Fig.~\ref{fig:framework}, we present the Class-wise Dynamic Graph Convolution Network to adaptively capture long-range contextual information. We use the ResNet-101 pretrained on the ImageNet dataset as the backbone, replace the last two down-sampling operations and employ dilation convolutions in the subsequent convolutional layers, hence enlarging the resolution and receptive field of the feature map, so the output stride becomes 8 instead of 16.%, preserving more details. 

Our model consists of two parts: basic network and CDGC module. Specifically, we adopt ResNet-101 together with atrous spatial pyramid pooling(ASPP) as the basic complete segmentation network. An input image is passed through the backbone and ASPP module, then produces a feature map $X\in \mathbb{R}^{C\times H\times W}$, where $C$,$H$,$W$ represent channel number, height and width respectively. Then we apply a convolution layer to realize the dimension reduction and the feature $X$ will participate in two different branches. The first branch is the classification step which produces the coarse segmentation prediction map. After that, the prediction map is transformed into masks for different classes, the masks and the feature $X$ are subsequently fed into the CDGC module to perform class-wise graph reasoning. And the output feature of our CDGC module is concatenated with the input feature, and refined through a $1\times1 $ conv to get the final refined segmentation result.

\begin{figure*}[t]
    \centering
    \includegraphics[width= 1\textwidth]{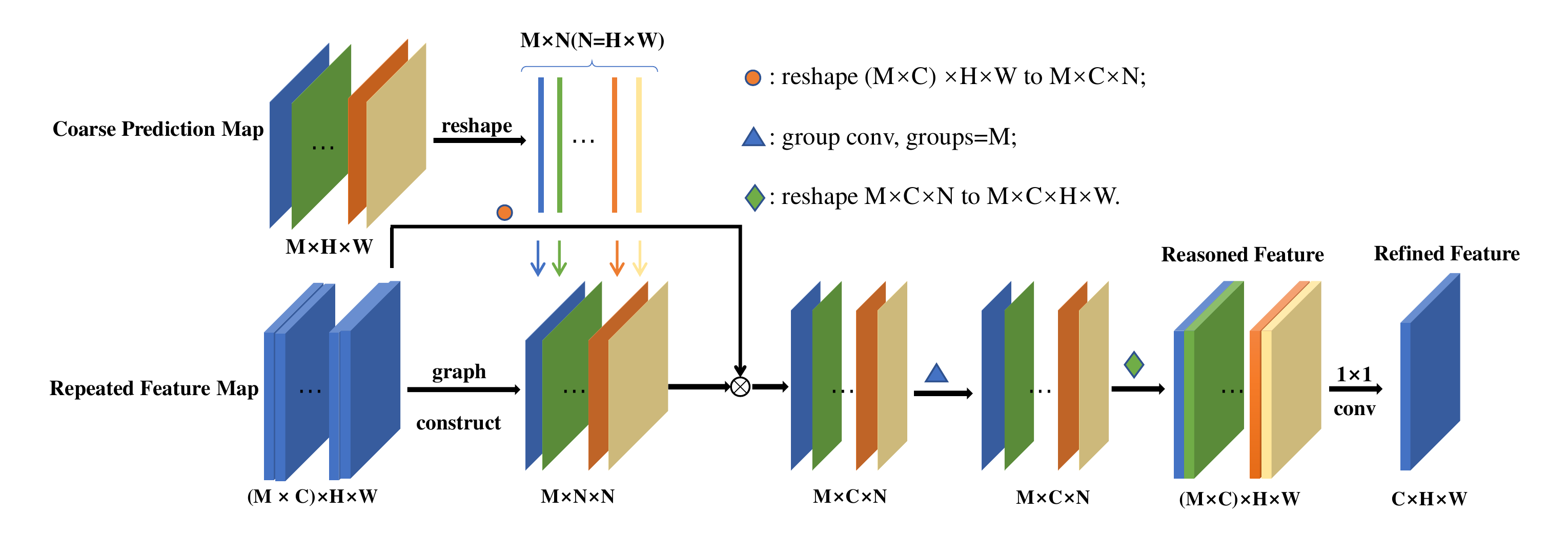}
    \caption{The details of Class-wise Dynamic Graph Convolution Module.}
    \label{fig:module}
\end{figure*}

\subsection{Class-wise Dynamic Graph Convolution Module}
The detailed structure of CDGC module is shown in Fig.~\ref{fig:module}. It consists of two subsequent processes, including graph construction and reasoning. The proposed module is based on a coarse-to-fine framework, where the input is the feature map $X$, coarse prediction map and the output is the refined feature map.
%\vspace{-10pt}
~\\

\noindent
\textbf{Class-wise Learning Strategy. }
%\vspace{-5pt}
Different from previous works~\cite{chen2019graph,liang2018symbolic,li2018beyond}  where graph construction is performed on all the nodes of different classes in the feature map, we adopt a class-wise learning strategy. There are several advantages. First, contextual information from different classes is considered separately so that the irrelevant region can be excluded to avoid the difficulty of learning. Second, it is easy to hard-mine the important information for a binary task (determine whether it is the target class or not) compared to the multi-class task learning. 

%To overcome the aforementioned issue, we develop a novel class-wised learning strategy where contextual information is acquired in class-oriented manner since the features of pixels belonging to different classes are not supposed to be fused together. 
Specifically, in the training process, a coarse-to-fine framework is adopted. The coarse prediction can be generated from a basic network. Each coarse predicted category is utilized to filter out the corresponding category and perform a graph construction based on the filtering operation. Hence, graph reasoning and information transmission only occur inside the chosen category, protecting the process of context aggregation from the interference of features in other categories.
%\vspace{-10pt}

\noindent
\textbf{Graph Construction. }
\noindent
\textbf{(1) Similarity Graph. } Intuitively, we can build the graph (which is adjacency matrix in our formulation) based on the similarity between different nodes, for two node features $x_i, x_j$, the pairwise similarity between two nodes is defined as,
\begin{equation}
    F(\bm{x_i,x_j})= \phi(\bm{x_i})^{T}\phi'(\bm{x_j}),
\end{equation}
where $\phi,\phi'$ denote two different transformations of the original features. In practice, we adopt linear transformations, hence $\phi(\bm{x})=\bm{wx}$ and $\phi'(\bm{x})=\bm{w'x}$. The parameters $\bm{w}$ and $\bm{w'}$ are both $D\times D$ dimensions weights which can be learned via back propagation, forming a dynamically learned graph construction method. After computing the similarity matrix, we perform normalization on each row of the matrix so that the sum of all the edge values connected to one node $i$ will be 1. In practice, we choose softmax as normalization function, so the output adjacency matrix will be,
\begin{equation}
    A_{ij}=\frac{exp(F(\bm{x_i,x_j}))}{\sum_{j=1}^N exp(F(\bm{x_i,x_j}))}
\end{equation}

\begin{figure}
    \centering
    \includegraphics[height=4.5cm]{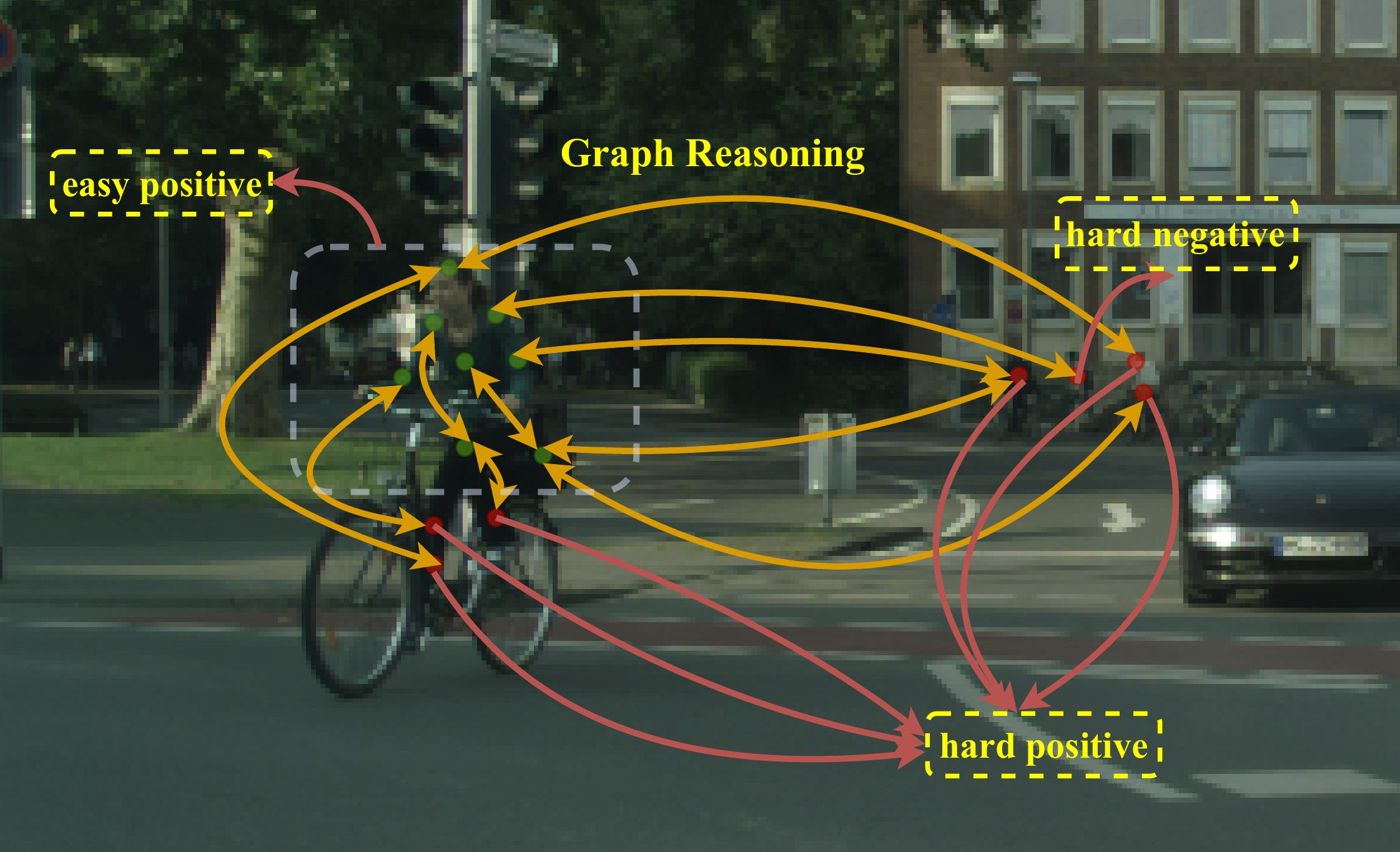}
    \caption{Illustration of dynamic sampling method. For one category 'rider' in this image, green and red points denote easy and hard samples, respectively. Hard positive samples consist of distant objects and boundaries. And hard negative samples denote the illegible object (person) in this image, which is likely to be recognized as rider.}
    \label{fig:ds}
\end{figure}

\noindent
\textbf{(2) Dynamic Sampling. } The original sampling method adopts a fully-connected fashion for pixels in the same category. However, since the prediction mask is obtained from a coarse segmentation result, it is possible that the sampled pixels are not actually belong to the same category, which makes the sampled set include `easy positive' part and `hard negative' part. In order to allocate enough attention to these hard-to-classify pixels, we develop a dynamic sampling method which focuses on selecting out these hard pixels. As shown in Fig.~\ref{fig:ds}, in the training process, we take coarse segmentation mask and groundtruth mask as input, and compute the intersection set between them, which is pure `easy positive' part. Formally, we denote the coarse segmentation mask, groundtruth mask set as $C$ and $G$ respectively, hence the intersection set can be denoted as $C\cap G$. Then with coarse segmentation mask subtracting the intersection set, the rest part is pure `hard negative' denoted as $C-C\cap G$. Similarly, with groundtruth mask getting rid of the intersection set, the rest part is pure `hard positive', denoted as $G-C\cap G$. Besides, some ratio of `easy positive' samples are needed to guide the learning of these hard pixels, so we randomly choose some ratio of pixels from the intersection set which consists of pure `easy positive' samples, so we finally get the sampled set denoted as,
%\vspace{-5pt}
\begin{equation}
  \begin{aligned}
      Sampled  &=C-C\cap G + G-C\cap G + ratio\cdot C\cap G \\ 
               &= C\cup G-(1-ratio)\cdot C\cap G
  \end{aligned}
\end{equation}
Therefore, with this dynamic sampling method, our graph construction process can pay enough attention to these hard pixels.

Specifically, dynamic sampling is only used at the training stage but not the inference stage. At the training stage, we use both coarse prediction mask and groundtruth mask to mine hard positive and negative samples in a class-wise manner. Besides, some easy positive samples are also selected to guide the hard samples learning. All these samples compose the graph nodes for the training stage. At the inference stage, pixels in the same category according to the coarse prediction mask are sampled to construct the graph. 

%\vspace{-8pt}
\noindent
\textbf{Graph Reasoning. }
Discriminative pixel-level feature representations are essential for semantic segmentation, which could be obtained by the proposed graph convolution based module in a class-wise manner. By exploiting the relations between pixels sampled by category, the intra-class consistency can be preserved and moreover, inter-class discrepancy can also be enhanced with our dynamic sampling method.

The detailed structure of CDGC module is shown in Fig.~\ref{fig:module}. The module takes the repeated feature map $\bm{X}\in \mathbb{R}^{(M\times C)\times H\times W}$ and coarse prediction map $P\in \mathbb{R}^{M\times H\times W}$ as input, where $M$, $C$, $H$, $W$ denote the number of classes in the dataset, dimension of the feature map, height and width, respectively. Inspired by point cloud segmentation~\cite{qi2017pointnet,wang2019dynamic}, we treat nodes in the feature map as the vertexes in the graph. Therefore, we transform the feature map to the graph representation: $X\in \mathbb{R}^{M\times C\times N}$, where $N=H\times W$ denotes the number of nodes in the feature map. Similarly, we transform the coarse prediction map into $P\in \mathbb{R}^{M\times N}$. Applying the graph construction methods discussed above, we can obtain the adjacency matrix of the feature map for each category, treating each graph feature $x\in \mathbb{R}^{C\times N}$ separately (M in total), thus producing M adjacency matrices integrated as $A\in \mathbb{R}^{M\times N\times N}$. 

Following the paradigm of graph convolution, we multiply the adjacency matrix and the transposed feature map to get the sampled feature map $X\in \mathbb{R}^{M\times C\times N} $.  Subsequently, group graph convolution is performed, resulting in a feature $X\in \mathbb{R}^{M\times C \times N}$ which will be reshaped back to the original grid form: $X\in \mathbb{R}^{M\times C \times H \times W }$.  Then a $1\times 1$ conv is performed to learn the weights of adaptively aggregrating feature maps for M classes, producing a refined feature $X\in \mathbb{R}^{C\times H\times W}$. Once obtaining the refined feature map, we combine this feature map with the input feature map to get the final output. Specifically, the combine method is concatenation or summation. 
Finally ,the output feature is passed through the conventional $1\times 1$ convolution layer to get the final segmentation prediction map.
%\vspace{-8pt}
\subsection{Loss Function}
Both coarse and refined output are supervised with the semantic labels. Moreover, following normal practice in previous state-of-the-art works~\cite{zhao2017pyramid,zhu2019asymmetric,zhang2019acfnet}, we add the auxiliary supervision for improving the performance, as well as making the network easier to optimize. Specifically, the output of the third stage of our backbone ResNet-101 is further fed into a auxiliary layer to produce a auxiliary prediction, which is supervised with the auxiliary loss. As for the main path, coarse segmentation result and refined segmentation result are produced and hence require proper supervision. We apply standard cross entropy loss to supervise the auxiliary output and the coarse prediction map, and employ OHEM loss~\cite{shrivastava2016training} for the refined prediction map. In a word, the loss can be formulated as follows,
\begin{equation}
    L=\alpha \cdot l_c+\beta \cdot l_f+\gamma \cdot l_a
\end{equation}
where $\alpha,\beta,\gamma $ are used to balance the coarse prediction loss $l_c$, refined prediction loss $l_f$ and auxiliary loss $l_a$.

\section{Experiments}
\label{exp}
%\vspace{-8pt}
To evaluate the performance of our proposed CDGC module, we carry out extensive experiments on benchmark datasets including Cityscapes~\cite{cordts2016cityscapes}, PASCAL VOC 2012~\cite{everingham2010pascal} and COCO Stuff~\cite{caesar2018coco}. Experimental results demonstrate that the proposed method can effectively boost the performance of the state-of-the-art methods. In the following section, we will introduce the datasets and implementation details, and then perform ablation study on Cityscapes dataset. Finally, we report the results on PASCAL VOC 2012 dataset and COCO Stuff dataset.
%\vspace{-10pt}
\subsection{Datasets and Evaluation Metrics}
%\vspace{-5pt}
\noindent
\textbf{Cityscapes.} The Cityscapes dataset~\cite{cordts2016cityscapes} is tasked for urban scene understanding, which contains 30 classes and only 19 classes of them are used for scene parsing evaluation. The dataset contains 5000 finely annotated images and 20000 coarsely annotated images. The finely annotated 5000 images are divided into 2975/500/1525 images for training, validation and testing.

\noindent
\textbf{PASCAL VOC 2012.} The PASCAL VOC 2012 dataset~\cite{everingham2010pascal} is one of the most competitive semantic segmentation dataset which contains 20 foreground object classes and 1 background class. The dataset is split into 1464/1449/1556 images for training, validation and testing. ~\cite{hariharan2011semantic} has augmented this dataset with annotations ,resulting in 10582 train-aug images.

\noindent
\textbf{COCO Stuff.} The COCO Stuff dataset~\cite{caesar2018coco} is a challenging scene parsing dataset containing 59 semantic classes and 1 background class. The training and test set consist of 9K and 1K images respectively.

\noindent
In our experiments, the mean of class-wise Intersection over Union (mIoU) is used as the evaluation metric.

\subsection{Implementation Details}
We choose the ImageNet pretrained ResNet-101 as our backbone and remove the last two down-sampling operations, and employ dilated convolutions in the subsequent convolution layers, making the output stride equal to 8. For training, we use the stochastic gradient descent(SGD) optimizer with initial learning rate 0.01, weight decay 0.0005 and momentum 0.9 for Cityscapes dataset. Moreover, we adopt the `poly' learning rate policy, where the initial learning rate is multiplied by $(1-\frac{iter}{max\_iter})^{power}$ with power=0.9. For Cityscapes dataset, we adopt the crop size as $769\times 769$, batch size as 8 and training iterations as 30K. For PASCAL VOC 2012 dataset, we set the initial learning rate as 0.001, weight decay as 0.0001, crop size as $513\times 513$, batch size as 16 and training iterations as 30K. For COCO Stuff dataset, we set initial learning rate as 0.001, weight decay as 0.0001, crop size as $520\times 520$, batch size as 16, and training iterations as 60K.
Moreover, the loss weights $\alpha, \beta, \gamma$ are set to be 0.6, 0.7 and 0.4 respectively.
\subsection{Ablation Study}
In this subsection, we conduct extensive ablation experiments on the validation set of Cityscapes with different settings for our proposed CDGCNet.

\noindent
\textbf{The impact of class-wise learning strategy.} We use the dilated ResNet-101 as the baseline network, and final segmentation result is obtained by directly upsampling the output. To evaluate the effectiveness of the proposed  class-wise learning strategy, we carry out the experiments where plain GCN and class-wise GCN are adopted separately. Concretely, plain GCN is realized by simply performing graph construction operation on the feature map obtained from the backbone, while class-wise GCN is realized in a class-wise manner. Their graph construction methods are similar. As shown in Table~\ref{tab:class}, the proposed class-wise GCN reasoning performs better than the plain GCN. Since plain-CGN adopts fully connected fashion onto the input feature map, it serves similarly as self-attention based method, which is likely to mislead the contextual information aggregation with features from pixels of other categories, while our method, on the other hand, is capable of avoiding this kind of problem.

\begin{table}[t]
\begin{minipage}[t]{0.41\textwidth}
\renewcommand\arraystretch{0.7}
\begin{center}
\caption{Performance comparisons of our proposed class-wise GCN and plain-GCN on Cityscapes validation set.}
\begin{tabularx}{5cm}{p{3.9cm}|X<{\centering}}
\toprule[1.5pt]
Method & mIOU\\ & (\%) \\
\midrule[1pt]
\midrule[1pt]
ResNet-101 Baseline & 76.3 \\
ResNet-101 + plain-GCN & 78.2 \\
ResNet-101 + class-GCN & 79.4 \\
\bottomrule[1.5pt]
\end{tabularx}
\label{tab:class}
\end{center}
\end{minipage}
%\end{table}
%\begin{table}
\quad
\begin{minipage}[t]{0.54\textwidth}
\renewcommand\arraystretch{0.7}
\begin{center}
\caption{Detailed performance comparisons of our proposed Class-wise Dynamic Graph Convolution module on Cityscapes validation set.}
\begin{tabularx}{6.9cm}{p{5.9cm}|X<{\centering}}
\toprule[1.5pt]
Method & mIOU \\ & (\%)\\
\midrule[1pt]
\midrule[1pt]
ResNet-101 Baseline & 76.3 \\
ResNet-101 + ASPP & 78.4 \\
ResNet-101 + CDGC(concat) & 79.4 \\
ResNet-101 + CDGC(sum) & 79.2 \\
ResNet-101 + ASPP + CDGC(sum) & 79.9 \\
ResNet-101 + ASPP + CDGC(concat) & 80.0\\
\bottomrule[1.5pt]
\end{tabularx}
\label{tab:module}
\end{center}
\end{minipage}
\end{table}

\begin{figure*}
    \centering
    \includegraphics[width=1\textwidth]{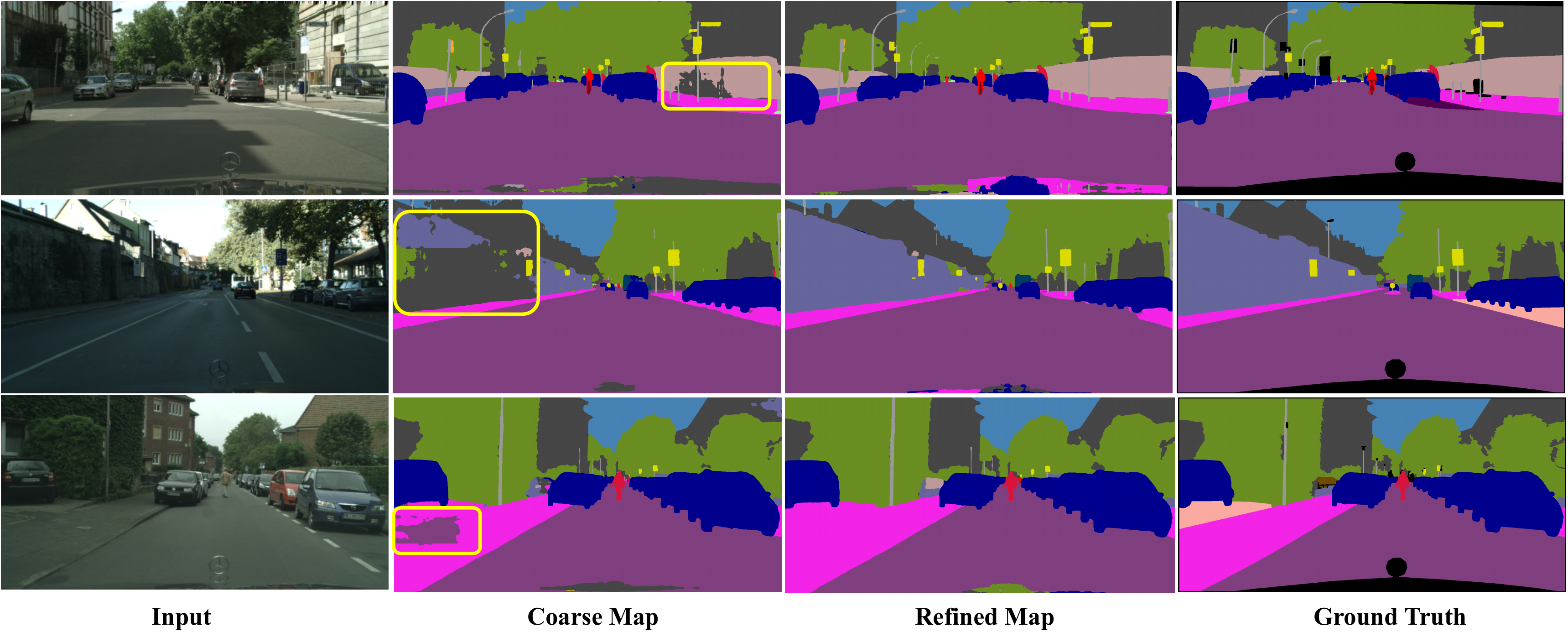}
    \caption{Visualization results on Cityscapes validation set.}
    \label{fig:visualize}
\end{figure*}

\noindent
\textbf{The impact of CDGC module.} 
Based on the dilated ResNet-101 backbone, we subsequently add ASPP module and the proposed module to evaluate the performance, as shown in Table~\ref{tab:module}. The graph is constructed based on dynamic similarity. The result of solely adding ASPP module is 78.4\%, which is about 1\% lower than solely adding CDGC module. Furthermore, we perform experiments on the feature aggregation manners which include concatenation and summation. As results shown in Table~\ref{tab:module}, the CDGC module can significantly improve the performance over the baseline network by 3\% in mIOU and concatenation method is slightly better than the summation one, so we will use concatenation aggregation method in later comparisons. Finally, we choose ResNet-101 plus ASPP module as our basic segmentation network and use CDGC module to get the final refined prediction map, achieving 1.6\% gain in mIOU, which demonstrates that CDGC module can be easily plugged into any state-of-the-art segmentation network to further boost the performance. The effect of CDGC module can be visualized in Fig.~\ref{fig:visualize}. Some details and boundaries are refined compared to the coarse map predicted by the basic network. These results prove that our proposed CDGC module can significantly capture long-range contextual information together with local cue and also preserve intra-class consistency, which can effectively boost the performance of segmentation.
%\vspace{-1cm}

\noindent
\textbf{Comparisons of different graph construction methods. }
In this subsection, we evaluate the performance of our module using two different graph construction methods mentioned before. Specifically, we use ResNet-101+ASPP as basic segmentation network and the original feature is concatenated with refined feature to get the final prediction map. Table~\ref{tab:construction} indicates the performance on Cityscapes validation set by adopting different graph construction method,
where `sim' denotes the similarity graph method and `ds' denotes the dynamic sampling method and the easy positive sampling ratio is set as [0.2, 0.4, 0.6, 0.8]. As can be seen in Table~\ref{tab:construction}, 
as the easy positive sampling ratio grows, the performance becomes better since the easy positive samples serve as the guiding criterion for learning the reasonable weights for hard samples. From the result shown in Table~\ref{tab:construction}, when sampling ratio is above 0.4, the dynamic sampling method can outperform the similarity graph method since it gives more attention to hard samples including hard positive ones and hard negative ones while similarity graph adaptively learn the parameters of the construction weights, which may not be efficiently learned in similarity graph method.
\begin{table}[t]
\begin{minipage}[t]{0.5\textwidth}
\renewcommand\arraystretch{0.7}
\begin{center}
\caption{Performance comparisons of graph construction method on Cityscapes validation set.}
\begin{tabularx}{6.9cm}{p{5.8cm}|X<{\centering}}
\toprule[1.5pt]
Method & mIOU\\ & (\%) \\
\midrule[1pt]
\midrule[1pt]
ResNet-101 + ASPP & 78.4 \\
ResNet-101 + ASPP + CDGC(sim) & 80.0\\
ResNet-101 + ASPP + CDGC(ds 0.2) & 79.8  \\
ResNet-101 + ASPP + CDGC(ds 0.4) & 80.3\\
ResNet-101 + ASPP + CDGC(ds 0.6) & 80.8\\
ResNet-101 + ASPP + CDGC(ds 0.8) & 80.9 \\
ResNet-101 + ASPP + CDGC(ds 1.0) & 81.1 \\

\bottomrule[1.5pt]
\end{tabularx}
\label{tab:construction}
\end{center}
\end{minipage}
%\end{table}
%\begin{table}
%\begin{comment}
\qquad
\quad
\begin{minipage}[t]{0.4\textwidth}
\renewcommand\arraystretch{0.5}
\begin{center}
\caption{Performance influences with different evaluation strategies on Cityscapes validation set.}
~\\
\begin{tabularx}{4.7cm}{p{1.6cm}|X<{\centering}|X<{\centering}|X<{\centering}}
\toprule[1.5pt]
Method & MS & Flip & mIOU\\ & & & (\%) \\
\midrule[1pt]
\midrule[1pt]
CDGCNet &  &  & 81.1 \\
CDGCNet  & \checkmark & & 81.6 \\
CDGCNet    & & \checkmark & 81.4 \\
CDGCNet     & \checkmark & \checkmark & \textbf{81.9} \\
\bottomrule[1.5pt]
\end{tabularx}
\label{tab:trick}
\end{center}

\end{minipage}
\end{table}

\noindent
\textbf{The impact of hard samples. }
We further perform experiments to evaluate the impact of hard samples utilized in dynamic sampling method. At the training stage, we construct the graph with dynamic sampling method while keeping the ratio of easy positive samples as 1.0. From the result shown in Table \ref{tab:hard}, utilizing hard samples can improve the performance since extra attention can be paid to hard pixels, hence performing a better feature learning process.
\begin{table}
\renewcommand\arraystretch{0.7}
\caption{Performance comparisons of different samples used in dynamic sampling method on Cityscapes validation set.}
\begin{center}
\begin{tabularx}{9.3cm}{p{7.6cm}|X<{\centering}}
\toprule[1.5pt]
Sample  & mIOU(\%) \\
\midrule[1pt]
\midrule[1pt]
Easy Positive & 79.9 \\
Easy Positive + Hard Positive  & 80.5 \\
Easy Positive + Hard Negative  & 80.0 \\
Easy  Positive + Hard Positive + Hard Negative  & 81.1 \\
\bottomrule[1.5pt]
\end{tabularx}
\end{center}

\label{tab:hard}
\end{table}
%\vspace{-1cm}
%\end{comment}

\noindent
\textbf{The impact of evaluation strategies. }
Based on details discussed above, we propose Class-wise Dynamic Graph Convolution Network (CDGCNet) with ResNet-101+ASPP as basic network and dynamic sampling method to construct the graph. Like previous work \cite{zhao2017pyramid,yang2018denseaspp,fu2019dual,huang2018ccnet,yuan2018ocnet}, we also adopt the left-right flipping and multi-scale [0.75, 1.0, 1.25, 1.5, 1.75, 2.0] evaluation strategies. From Table~\ref{tab:trick}, MS/Flip improves the performance by 0.8\% on validation set.

\begin{figure}[!h]
    \centering
    \includegraphics[width=1\textwidth]{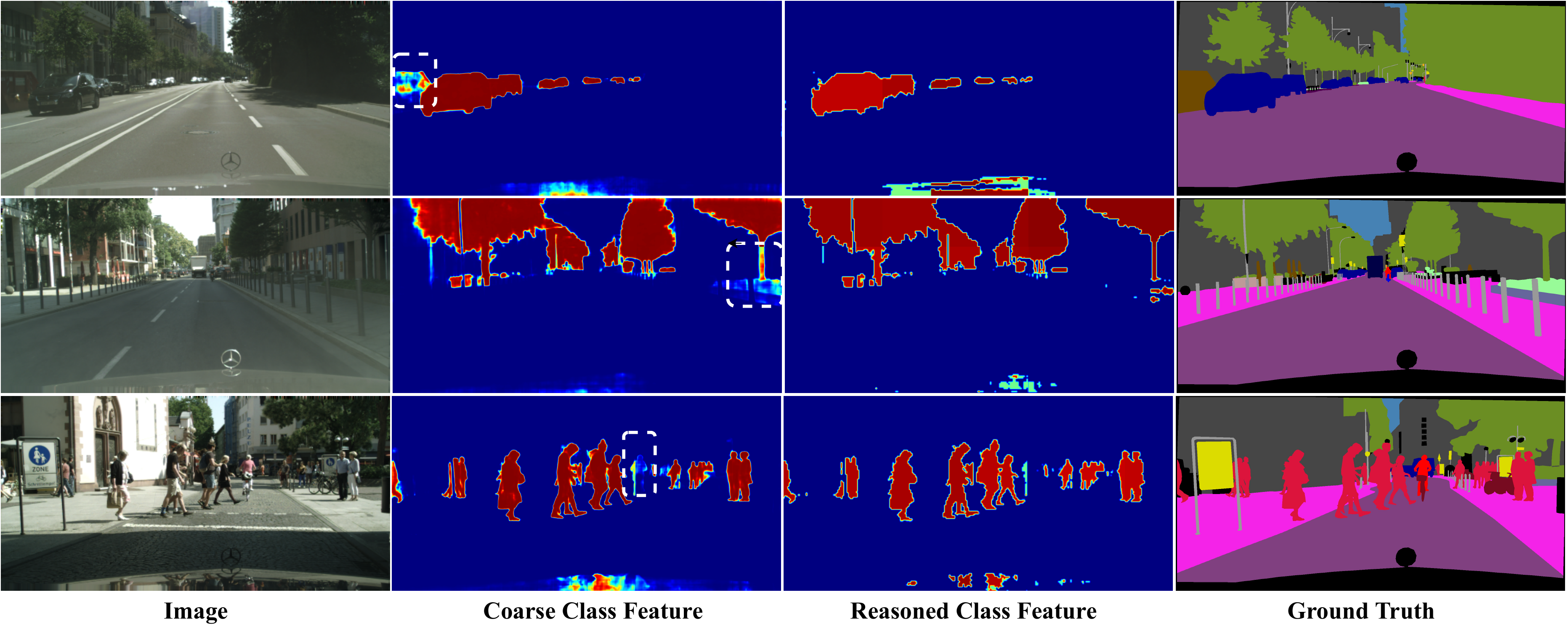}
    \caption{Visualizations of class-wise features before and after graph convolution on Cityscapes validation set. From left to right: input image, class-wise feature before CDGC module, class-wise feature after CDGC module, ground truth. From top to bottom, the visualized category is car, vegetation and person.}
    \label{class-wise}
\end{figure}

%\vspace{-1cm}
\noindent
\textbf{Visualizations of class-wise features. }
Qualitative results are provided in Figure \ref{class-wise} to compare the difference of class-wise features before and after CDGC module. We use white squares to mark the challenging regions which compose of hard samples% including hard positive and hard negative pixels
. As shown in the figure, after class-wise dynamic graph convoluton, hard pixels can be effectively resolved. In particular, in the first and third lines, hard pixels are specified to hard negative pixels and can be successfully distinguished. While in the second line, hard pixels are specified to hard positive pixels, as shown in the visualization, ambiguity is well taken care of. Moreover, with dynamic sampling method mining hard samples, boundary information is preserved and enhanced, hence producing better results.

\begin{comment}
\noindent
\textbf{Impact on boundary accuracy. } To validate the effectiveness of dynamic sampling method in mining hard samples which occur in boundary areas, we design an experiment to show that our method improves boundary precision. Following \cite{kohli2009robust,marin2019efficient}, we adopt a standard trimap approach where we compute the classification accuracy within a band (called trimap) of varying width around boundaries of segments(shown in Figure \ref{boundary_visual}). As shown in Figure \ref{boundary_gain}, we use deeplab v3 as baseline and compute the boundary accuracy gain over baseline of CDGC module with two graph construction methods: similarity graph and dynamic sampling. The results indicate that dynamic sampling graph construction method effectively improves the vicinity of boundaries compared with baseline method and CDGC module with simple similarity graph, hence preserving more details.
\begin{figure}
\begin{minipage}[t]{0.495\textwidth}
    \centering
    \includegraphics[height=3.5cm]{figures/gain.pdf}
    \caption{Boundary accuracy gain over baseline for two graph construction methods on Cityscapes validation set.}
    \label{boundary_gain}
    \end{minipage}
\hfill
\begin{minipage}[t]{0.495\textwidth}
    \centering
    \includegraphics[height=3.5cm]{figures/boundary_visual.pdf}
    \caption{Trimaps used for boundary accuracy evaluation with different band width from Cityscapes validation set}
    \label{boundary_visual}
\end{minipage}
\end{figure}
\end{comment}
%\vspace{-1 cm}

\subsection{Comparisons with state-of-the-arts}
Furthermore, we evaluate our method on the test set of three benchmark datasets: Cityscapes, PASCAL VOC 2012 and COCO Stuff datasets. Specifically, we use ResNet-101 as backbone, dynamic sampling method with ratio 1.0 as graph construction method. Moreover, we train the proposed CDGCNet with both training and validation set and use the multi-scale and flip strategies while testing. From Table~\ref{all_result}, it can be observed that our CDGCNet achieves state-of-the-art performance on all three benchmark datasets.
\begin{table*}[t]%\small
%\begin{spacing}{0.8}
\caption{Comparisons with State-of-the-art methods on three benchmark datatsets.}
\begin{center}

\scalebox{1}{ 
\setlength{\tabcolsep}{1mm}{
\begin{tabularx}{11.5cm}{p{3cm}|p{2.3cm} |X<{\centering} |X<{\centering}|X<{\centering}}
\toprule[1.5pt]
&  & Cityscapes & PASCAL VOC 2012 & COCO Stuff \\
Methods & Backbone & mIOU($\%$) & mIOU($\%$)& mIOU($\%$) \\
\midrule[1pt]

FCN~\cite{long2015fully} & VGG-16 & - & 62.2 & 22.7 \\
DeepLab-CRF~\cite{chen2017deeplab} & VGG-16 & - & 71.6 & - \\
DAG-RNN~\cite{shuai2017scene} & VGG-16 & - & - & 31.2\\
RefineNet~\cite{lin2017refinenet} & ResNet-101 & 73.6 & - & 33.6 \\
GCN~\cite{peng2017large} &ResNet-101 &76.9& - & -\\
SAC~\cite{zhang2017scale} & ResNet-101 & 78.1& - & -\\
CCL~\cite{ding2018context} & ResNet-101 & - & - & 35.7\\
PSPNet~\cite{zhao2017pyramid} & ResNet-101 & 78.4 & 82.6 & -\\
BiSeNet~\cite{yu2018bisenet} & ResNet-101 & 78.9 & - & -\\
DFN~\cite{yu2018learning} & ResNet-101 & 79.3 & 82.7 &  -\\
DSSPN~\cite{liang2018dynamic} & ResNet-101 & - & - & 37.3 \\
SGR~\cite{liang2018symbolic} & ResNet-101 & - & - & 39.1 \\
PSANet~\cite{zhao2018psanet} & ResNet-101 & 80.1 & - & -   \\
DenseASPP~\cite{yang2018denseaspp} & DenseNet-161 & 80.6 & - & - \\
GloRe~\cite{chen2019graph} & ResNet-101 & 80.9 & - & - \\
EncNet~\cite{zhang2018context} & ResNet-101 & - & 82.9 & - \\
DANet~\cite{fu2019dual} & ResNet-101 & 81.5 & 82.6 & 39.7\\
\midrule[1pt]
{\bf {CDGCNet(Ours)}} & ResNet-101 & \textbf{82.0} & \textbf{83.9} & \textbf{40.7}\\
\bottomrule[1.5pt]
\end{tabularx}}}
\\
\end{center}
\label{all_result}
%Our Se-ResNet50 version MFT and ResNet50 version RCO achieve the first place and third place respectively.
%\end{spacing}

\end{table*} 
\section{Conclusions}

In this paper, we have presented the Class-wise Dynamic Graph Convolution Network (CDGCNet) which can adaptively capture long-range contextual information, hence performing a reliable graph reasoning along nodes for better feature aggregation and weight allocation. Specifically, we utilize a class-wise learning strategy to enhance contextual learning. Moreover, we develop a dynamic sampling method for graph construction, which gives extra attention to hard samples, thus benefiting the feature learning. The ablation experiments demonstrate the effectiveness of CDGC module. Our CDGCNet achieves outstanding performance on three benchmark datasets, \textit{i.e.}  Cityscapes, PASCAL VOC 2012 and COCO Stuff.

%\clearpage\mbox{}Page \thepage\ of the manuscript.
%\clearpage\mbox{}Page \thepage\ of the manuscript.

%This is the last page of the manuscript.
%\par\vfill\par
%Now we have reached the maximum size of the ECCV 2020 submission (excluding references).
%References should start immediately after the main text, but can continue on p.15 if needed.

\clearpage
%\input{2638.bbl}
\bibliographystyle{splncs04}
\bibliography{2638}
\end{document}